# Two-Stage Swarm Intelligence Ensemble Deep Transfer Learning (SI-EDTL) for Vehicle Detection Using Unmanned Aerial Vehicles

Zeinab Ghasemi Darehnaei [1], Mohammad Shokouhifar [2,3,*], Hossein Yazdanjouei [4,5], S.M.J. Rastegar Fatemi [1]

[1] *Department of Electrical Engineering, College of Engineering, Saveh Branch, Islamic Azad University, Saveh, Iran*
[2] *Department of Electrical and Computer Engineering, Shahid Beheshti University, Tehran, Iran*
[3] *Institute of Research and Development, Duy Tan University, Da Nang 550000, Vietnam*
[4] *Microelectronics Research Laboratory, Urmia University, Urmia, Iran*
[5] *Department of Computer Science, Khazar University, Baku AZ1096, Azerbaijan*

*Abstract*: This paper introduces SI-EDTL, a two-stage swarm intelligence ensemble deep transfer learning model for detecting multiple vehicles in UAV images. It combines three pre-trained Faster R-CNN feature extractor models (InceptionV3, ResNet50, GoogLeNet) with five transfer classifiers (KNN, SVM, MLP, C4.5, Naïve Bayes), resulting in 15 different base learners. These are aggregated via weighted averaging to classify regions as Car, Van, Truck, Bus, or background. Hyperparameters are optimized with the whale optimization algorithm to balance accuracy, precision, and recall. Implemented in MATLAB R2020b with parallel processing, SI-EDTL outperforms existing methods on the AU-AIR UAV dataset.

**\* Correspondence:** Mohammad Shokouhifar

Email: mohammadshokouhifar@duytan.edu.vn, ORCID: 0000-0001-7370-4760.

**Email addresses:**

Zeinab Ghasemi Darehnaei: zeiynab.ghaasemi@gmail.com
Mohammad Shokouhifar: mohammadshokouhifar@duytan.edu.vn
Hossein Yazdanjouei: h.yazdanjouei@gmail.com
S.M.J. Rastegar Fatemi: mohammadjalalrastegarfatemi@gmail.com

# 1. INTRODUCTION

Unmanned aerial vehicles (UAVs) are cost-effective tools widely used in remote sensing, particularly for capturing traffic data [1-3]. Previously, high costs and challenges in processing high-resolution images limited their use in transportation studies. However, affordable UAVs now support traffic monitoring, planning, and safety, offering advantages over ground-based sensors such as wide coverage, mobility, quick on-demand imaging, and minimal interference [4,5]. Beyond traffic, UAVs are applied in vegetation monitoring [6], urban management [7], disaster response, pipeline surveillance, archaeological exploration [8], and object detection [9-12].

Vehicle detection from UAV images has become a key research focus for traffic estimation, urban planning, and parking management [3,10-18]. Yet, it remains challenging due to changing illumination, complex backgrounds, and diverse traffic conditions [19]. Classical and machine learning methods often rely on handcrafted features, but deep learning, particularly convolutional neural networks (CNNs), has recently outperformed these approaches by automatically learning features [20-23]. Techniques like sliding-window CNNs [18] improved multi-scale detection but were computationally intensive, leading to advances such as SPPnet [24], R-CNN [25], Fast R-CNN [26], and Faster R-CNN [27], which balances accuracy and speed.

Motivated by Faster R-CNN's success, this paper introduces SI-EDTL, a swarm intelligence ensemble deep transfer learning model for multi-vehicle detection in UAV images. SI-EDTL combines three pre-trained Faster R-CNN feature extractors (InceptionV3, ResNet50, GoogLeNet) with five classifiers (KNN, SVM, MLP, C4.5, Naïve Bayes) to produce 15 base learners. Outputs are aggregated via weighted averaging to classify regions into four vehicle types or background. Hyperparameters, including ensemble weights and decision thresholds, are optimized using the whale optimization algorithm (WOA) to balance accuracy, precision, and recall. The model is tested on the AU-AIR UAV dataset for Car, Van, Truck, and Bus detection. Our key contributions can be summarized as

- SI-EDTL introduces a tunable, accurate deep transfer learning framework for multi-object detection.
- The SI-EDTL model uses a two-level ensemble of three Faster R-CNN feature extractors and five classifiers (3×5=15 base learners).
- The SI-EDTL model employs transfer learning on pre-trained deep models (InceptionV3, ResNet50, GoogLeNet) with five classifiers (KNN, SVM, MLP, C4.5, Naïve Bayes).
- The hyperparameters of the SI-EDTL modelare optimized using WOA to achieve the desired trade-off between accuracy, precision, and recall.
- The proposed SI-EDTL model demonstrated effectiveness on the AU-AIR UAV dataset for multiple vehicle detection.

The paper is organized as follows: Section 2 reviews existing vehicle detection methods; Section 3 introduces the SI-EDTL model; Section 4 presents experimental results; Section 5 concludes with future directions.

## 2. LITERATURE REVIEW

Over the years, extensive research has been conducted on vehicle detection. Traditional methods, such as background subtraction [28], frame difference [29], and optical flow [30], generally achieve low accuracy and often detect only moving vehicles. Classical algorithms like Viola-Jones (VJ) [31] and Discriminatively Trained Part-Based Models (DPM) [32] are more robust under noise and complex conditions, but they struggle with rotated objects, specific orientations, and illumination changes, often leading to high false-negative rates.

Machine learning approaches have been widely explored for vehicle detection in UAV images. For instance, Zhao and Nevatia [13] employed shadow and color intensities with a Bayesian classifier, while Kluckner et al. [15] used online boosting with LBP, Haar-like features, and orientation histograms. Other approaches include supervised sliding-window search for parked cars [16], combined features of LBP and HOG [17], and SIFT-based interest point extraction with SVM classification [11,13,33]. These methods handle high-resolution UAV images but often require complex feature engineering and computationally intensive steps.

Deep learning techniques have further advanced vehicle detection. Sliding-window CNNs [34] and hybrid CNN architectures [21] improve feature learning but suffer from high computational cost. R-CNN [25] introduced region proposals for object detection, though generating these regions is time-consuming. Variants like three-stage VGG16-based methods [35-37], convolutional SVM networks (CSVM) [38], and fully convolutional SVMs (FCSVM) [39] attempt to reduce computational cost while improving accuracy, but often involve complex multi-stage training or extensive processing.

To address these challenges, Faster R-CNN [27] was proposed as a more efficient solution for region proposal generation. Building on its success in both accuracy and speed, numerous techniques based on Faster R-CNN have been applied to vehicle detection from low-altitude UAV images [6,38-47]. Motivated by these achievements, this paper adopts the Faster R-CNN framework to detect vehicles efficiently and accurately from UAV imagery.

## 3. PROPOSED SI-EDTL MODEL

This paper presents SI-EDTL, a tunable swarm intelligence ensemble deep transfer learning model based on Faster R-CNN for multi-vehicle detection in UAV images. Faster R-CNN consists of two components: the Region Proposal Network (RPN), which generates candidate object regions of varying scales and aspect ratios, and the Fast R-CNN detector, which processes the entire image with CNN layers and refines the proposals. Sharing convolutional layers between RPN and Fast R-CNN enables high-quality object proposals with deep feature extraction. In the Fast R-CNN detector, each region of interest (RoI) is converted into a fixed-length feature vector via RoI pooling and passed through fully connected layers. The final outputs, computed through softmax and bounding-box regression, provide class scores for four vehicle types (Car, Van, Truck, Bus) and background, along with precise bounding boxes for each detected object.

## 3.1. Transfer learning model

In the proposed SI-EDTL model, three pre-trained CNNs (InceptionV3, ResNet50, and GoogLeNet) are transformed into Faster R-CNNs for feature extraction. This transfer learning process involves three steps: (1) replacing the final classification layers with new classifiers (KNN, SVM, MLP, C4.5, and Naïve Bayes) tailored to four vehicle classes (Car, Van, Truck, Bus); (2) adding a bounding box regression layer to refine object localization; and (3) connecting an RPN and RoI pooling layer to a suitable intermediate feature layer ("mixed7" in InceptionV3, "activation40_relu" in ResNet50, and "inception_4d-output" in GoogLeNet). Only the new layers are trained on the AU-AIR dataset, while the original convolutional weights remain unchanged.

The overall framework, illustrated in Fig. 1, shows how a pre-trained CNN is converted into a Faster R-CNN model. During transfer learning, the newly added modules are trained on UAV images, while the pre-trained convolutional base remains fixed. In the testing phase, the trained Faster R-CNN accurately detects multiple vehicle classes in unseen UAV images.

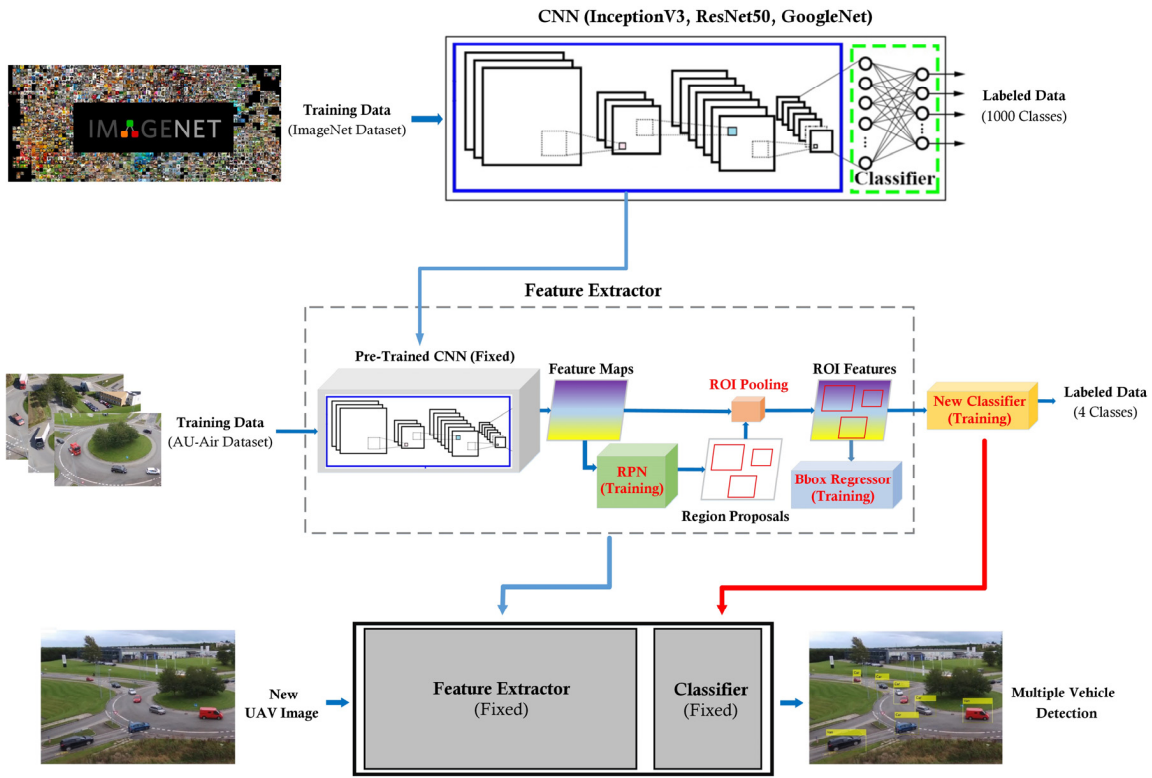

Fig. 1. Framework of transfer learning for a single base learner [48].

## 3.2. Multiple vehicle detection using SI-EDTL

The flowchart of training and test stages of the proposed multiple vehicle detection method using the SI-EDTL model can be seen in Fig. 2. At the training stage, a number of $N_1^{Train}$ cropped target objects of four object classes of multiple "vehicles" (Car, Van, Truck, and Bus) in different orientations, as well as a number of $N_0^{Train}$ of "background" class, which have been extracted from different UAV images, are used to train each base learner. These cropped images include label 0 for background (no-vehicle), and labels 1 to 4 for object classes Car, Van, Truck, and Bus, respectively. These cropped images are used as the same for training of $N_{BL}$ transfer learners within $N_{FE} \times N_{CL}$ ensemble structure of the SI-EDTL model. After training of different base learners and hyperparameter tuning of the SI-EDTL via K-fold cross validation method on the training dataset using WOA, the trained models as well as the optimized parameters of the SI-EDTL model are stored.

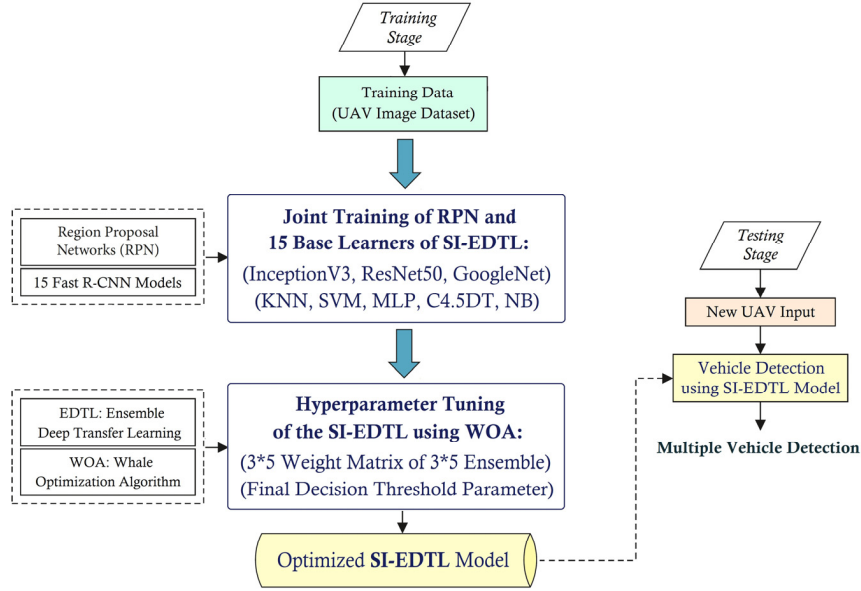

Fig. 2. Flowchart of multiple vehicle detection by the SI-EDTL model [48].

## 3.3. Ensemble learning model

Multiple vehicle detection in the SI-EDTL model is done using an ensemble of deep transfer learners in two stages, as seen in Fig. 3. The first stage includes Faster R-CNN feature extractors, which have been pre-trained on ImageNet dataset. The second stage includes classifiers, used for transfer learning of each pre-trained CNN model. As a result, $N_{BL}=N_{FE} \times N_{CL}=15$ base learners are obtained, where $N_{FE}=3$ and $N_{CL}=5$. The final object score of the region proposal $k$ for class $c$ is calculated via a weighted averaging of the outputs achieved by two-stage base learners as:

$$Out_c^k = \frac{\sum_{j=1}^{N_{CL}} \sum_{i=1}^{N_{FE}} w_{ij} Out_{ijc}^k}{\sum_{j=1}^{N_{CL}} \sum_{i=1}^{N_{FE}} w_{ij}} \qquad (1)$$

where $N_{CL}$ is the number of classifiers, $N_{FE}$ is the number of feature extractors, and $Out_{ijc}^k$ is the binary predicted label of class $c$ via $j$-th classifier utilizing $i$-th feature extractor for region proposal $k$: $Out_{ijc}^k=1$, if $j$-th classifier of $i$-th feature extractor has predicted label $c$ for the region proposal $k$, otherwise, $Out_{ijc}^k=0$. Moreover, $w_{ij}$ is the weight of classifier $j$ for feature extractor $i$ within the ensemble model. After calculation of $Out_c^k$ of all classes, the region proposal $k$ belongs to the class with the highest $Out_c^k$, it $Out_c^k > DTh$, where $DTh$ is the decision threshold.

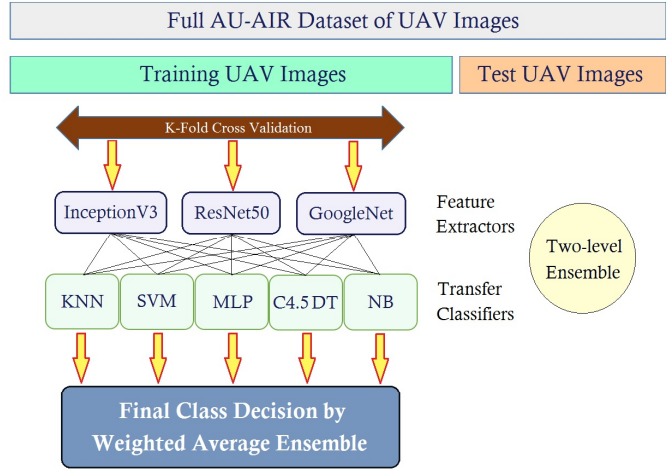

Fig. 3. Two-level ensemble structure of the SI-EDTL model [48].

### 3.4. Metaheuristic-driven Hyperparameter Tuning of SI-EDTL

Metaheuristics are high-level optimization strategies designed to efficiently search large and complex solution spaces where exact methods are impractical [49]. Their main strength lies in balancing exploration and exploitation, making them flexible, scalable, and adaptable to a wide range of real-world problems [50-52]. Over the past years, metaheuristics have been successfully applied across engineering, computer science, and data-driven domains [53-56]. Among them, the Whale Optimization Algorithm (WOA) [57] stands out as an effective choice for hyperparameter tuning, thanks to its simplicity, strong global search ability, and competitive performance in achieving well-balanced solutions. It simulates the hunting behavior of humpback whales by encircling the prey and update their position towards the best solution during execution of the algorithm [58].

As mentioned above, hyperparameters of the SI-EDTL are optimized using WOA once before applying the SI-EDTL model for online vehicle detection. To increase generalizability of the SI-EDTL model, K-fold cross validation with $K=10$ is applied for hyperparameter tuning on training dataset. The hyperparameters of the SI-EDTL model include the weights of the two-level $N_{FE} \times N_{CL}$ base learners and the final decision threshold parameter $DTh$. A solution (whale) for the optimization of the SI-EDTL model, Sol, can be encoded as a continuous matrix of dimension $N_{FE} \times N_{CL}$ and a continuous number representing $DTh$ parameter. According to Eq. (2), $i$-th row at $j$-th column of Sol.$W$ has a continuous value in the range [0,1], representing the weight of classifier $j$ for feature extractor $i$. Moreover, according to Eq. (3), Sol.$DTh$ can take a continuous value between 0 and 1, i.e., $DTh \in [0,1]$.

$$Sol.W(i,j) = w_{ij} \in [0,1]; \quad \forall i \in \{1,2,...,N_{FE}\}, j \in \{1,2,...,N_{TC}\} \tag{2}$$

$$Sol.DTh = DTh \in [0,1] \tag{3}$$

To evaluate each feasible solution, the solution is decoded to construct the SI-EDTL model, and then, its fitness value is evaluated as a weighted average function as:

$$\text{Fit} = W_P \times Precision_{avg} + W_R \times Recall_{avg} + W_A \times Accuracy \tag{4}$$

where $Precision_{avg}$ and $Recall_{avg}$ are the average precision and recall for object classes of vehicles, and *Accuracy* is the total accuracy of all regions, which can be expressed as:

$$Accuracy = \frac{Number\ of\ Correctly\ Classified\ Regions}{Total\ Number\ of\ Training\ Regions} \tag{5}$$

$$Precision_{avg} = \sum_{c=1}^{C} \frac{TP_c}{TP_c + FP_c} \times \frac{N_c}{N_{Total}} \tag{6}$$

$$Recall_{avg} = \sum_{c=1}^{C} \frac{TP_c}{TP_c + FN_c} \times \frac{N_c}{N_{Total}} \tag{7}$$

where $TP_c$ denotes the number of true positives for class $c$ (correctly identified objects), $FP_c$ the number of false positives (regions incorrectly classified as class $c$), and $FN_c$ the number of false negatives (objects of class $c$ not detected). In addition, $C$ represents the total number of vehicle classes, $N_c$ the number of training samples labeled as class $c$, and $N_{Total}$ the overall number of training samples across the four vehicle classes.

## 4. EXPERIMENTAL RESULTS

To justify the performance of the SI-EDTL model, it is applied for multiple vehicle detection in a UAV dataset, named AU-AIR [59], which is accessible in *https://bozcani.github.io/auairdataset*. As mentioned above, the AU-AIR dataset contains 32,283 labeled images with eight object classes. The full dataset includes 12,875 regions with 7,970 vehicles and 4,905 backgrounds. We use hold-out method to split up the whole dataset into "train" dataset with 75% and "test" dataset with the remained 25% of regions. The train dataset is used to train the different Faster R-CNN models as well as for hyperparameter tuning using K-fold cross validation ($K$=10). The details of the utilized dataset can be summarized in Table 1.

To adjust the controllable parameters of WOA, different values have been tested, and eventually, the best ones are considered for experiments. This mechanism has been widely used in literature [60-62]. As a result, the parameters of the WOA were set as MaxIter=500, PopSize=50, and $b$=1. Moreover, the weights of the different performance measures within the multi-objective function of Eq. (4) were set as $w_A$=0.5, $w_P$=0.3, and $w_R$=0.2.

TABLE I. NUMBER OF OBJECTS WITHIN EACH CLASS IN FULL DATASE, TRAN DATASET, AND TEST DATASET [48].

| Object Label | Full Dataset | Train Dataset | Test Dataset |
| --- | --- | --- | --- |
| Background | 4,905 | 3,683 | 1,222 |
| Car | 4,479 | 3,364 | 1115 |
| Van | 780 | 569 | 211 |
| Truck | 2,629 | 1,982 | 647 |
| Bus | 82 | 59 | 23 |
| All labeled regions | 12,875 | 9,657 (75%) | 3,218 (25%) |

## 4.1. Results

The results of the SI-EDTL model for some test UAV images where all their vehicles have been identified correctly, are shown in Fig. 4. All vehicles in Fig. 4 have been correctly identified, i.e., FPR=FNR=0. The results demonstrate the effectiveness of the SI-EDTL model to efficiently detect multiple vehicles.

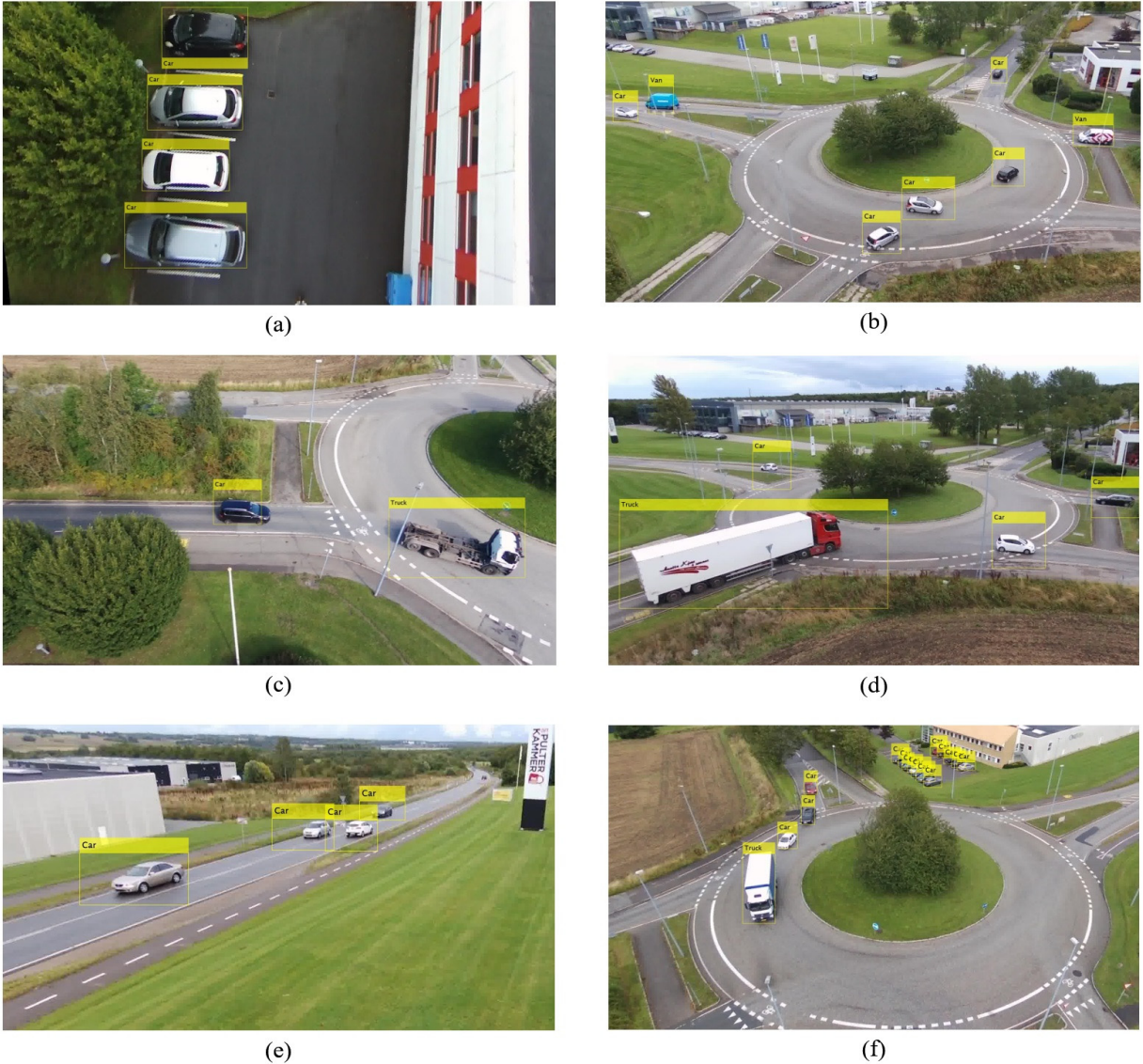

Fig. 4. Results of the proposed SI-EDTL model for some UAV images of AU-AIR test dataset [48].

## 4.2. Comparision with existing methods

To justify the performance of the proposed SI-EDTL, it is compared with a classical method (Viola-Jones) [31], a machine learning technique (HOG+SVM) [33], a deep convolutional neural network learning model (SW-CNN) [35] and two mobile object detectors YOLOv3-Tiny [63] and MobileNetv2-SSDLite [64]. Comparison of the proposed SI-EDTL model with the existing methods in terms of accuracy, precision, and recall, is summarized in Table 2.

As shown in Table 2, the proposed SI-EDTL model achieves superior detection performance compared to the YOLOv3-Tiny and MobileNetv2-SSDLite detectors. Utilizing lite mobile detectors to reach lower processing time and also not optimizing the default parameters for aerial images, explains the lower performance of these networks. Since in the proposed SI-EDTL model, we have developed a two-stage ensemble transfer learning comprising three sophisticated Faster R-CNNs at the first stage plus five transfer learning classifiers at the second stage, and also optimizing the hyperparameters of the model specifically for the AU-AIR dataset using WOA, better detection performance has been achieved.

TABLE II. COMPARISON OF SI-EDTL WITH EXISTING TECHNIQUES, IN TERMS OF ACCURACY, PRECISION, AND RECALL [48].

| Detection Method | Accuracy % | Precision % | Recall % |
| --- | --- | --- | --- |
| Viola-Jones [31] | 74.8 | 74.3 | 76.9 |
| HOG+SVM [33] | 77.5 | 84 | 74.4 |
| SW-CNN [35] | 83.5 | 88.1 | 83.7 |
| YOLOv3-Tiny [63] | 46.1 | 46.9 | 44.3 |
| MobileNetv2-SSDLite [64] | 55.3 | 58.7 | 53.5 |
| SI-EDTL (Proposed) | 91.3 | 89.3 | 89.1 |

## 4.3. Running time analysis

This section analyzes the running time of the proposed SI-EDTL, considering both offline training and online testing for vehicle detection in unseen UAV images. As summarized in Table 3, the training phase involves three stages: (i) training the Faster R-CNN feature extractors (InceptionV3, ResNet50, and GoogLeNet), (ii) applying transfer learning on five classifiers (KNN, SVM, MLP, C4.5DT, and NB), and (iii) tuning hyperparameters with WOA. In total, 15 classifiers are trained using the feature vectors extracted from the three models, and the overall offline training time largely depends on the number of Faster R-CNN feature extractors.

For online testing, parallel GPU processing is employed to run the three Faster R-CNN models concurrently. As a result, the test time is comparable to that of the slowest single extractor (InceptionV3), meaning the two-stage ensemble structure does not introduce significant additional cost. Figure 5 illustrates the required offline training time (in hours) and the online test time per UAV image (in seconds), highlighting the efficiency of SI-EDTL compared to single Faster R-CNN models.

TABLE III. COMPARISON OF RUNNING TIME OF SI-EDTL WITH DIFFERENT SINGLE FASTER R-CNN MODELS [48].

| Model | Offline Training Phase | | | Online Test for a UAV Image (seconds) |
| --- | --- | --- | --- | --- |
| | Faster R-CNN (hours) | Transfer Learning (seconds) | WOA (minutes) | |
| InceptionV3 | 9.6 | 15.1 | N/A | 1.43 |
| ResNet50 | 6.2 | 16.3 | N/A | 1.13 |
| GoogLeNet | 4.1 | 11.1 | N/A | 0.93 |
| SI-EDTL (Proposed) | 19.9 | 175 | 6.2 | 1.57 |

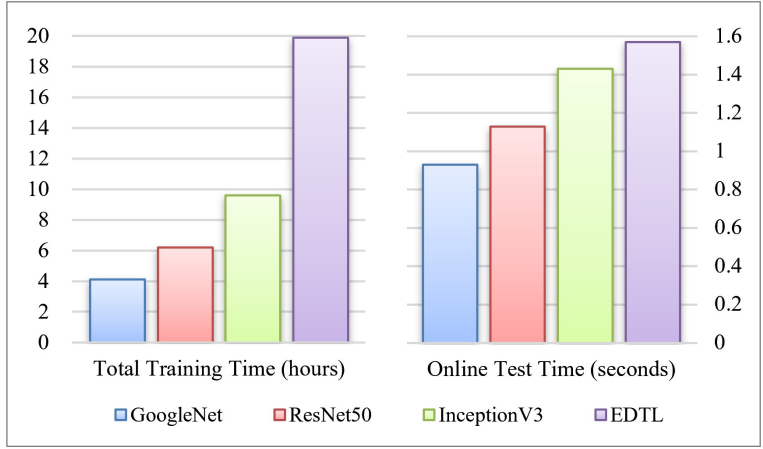

Fig. 5. Comparison of the total offline time and online test time of the SI-EDTL model with the single base models [48].

## 5. CONCLUSION

This paper introduced SI-EDTL, a swarm intelligence ensemble deep transfer learning model for multi-vehicle detection in UAV images. Built on Faster R-CNN, the model combines three pre-trained CNNs (InceptionV3, ResNet50, GoogLeNet) with five classifiers (KNN, SVM, MLP, C4.5, Naïve Bayes), forming 15 base learners aggregated through weighted averaging. Hyperparameters are tuned using the Whale Optimization Algorithm (WOA), allowing the model to balance accuracy, precision, and recall based on application needs.

Implemented with parallel GPU processing in MATLAB R2020b and tested on the AU-AIR dataset, SI-EDTL outperforms single Faster R-CNN models and other state-of-the-art techniques in accuracy, precision, recall, and mean average precision. Beyond its strong performance, the model is flexible and tunable, making it adaptable to different detection requirements. Future work will explore extending the ensemble with additional CNNs and classifiers, evaluating homogeneous ensembles through bagging, and training Faster R-CNNs from scratch with larger datasets and computing resources.